\documentclass[10pt]{article} 

\usepackage[accepted]{tmlr}

\usepackage{amsmath,amsfonts,bm}

\def\eqref#1{equation~\ref{#1}}

\def\1{\bm{1}}

\DeclareMathAlphabet{\mathsfit}{\encodingdefault}{\sfdefault}{m}{sl}
\SetMathAlphabet{\mathsfit}{bold}{\encodingdefault}{\sfdefault}{bx}{n}

\usepackage{hyperref}
\usepackage{url}

\usepackage{times}
\usepackage{epsfig}
\usepackage{graphicx}
\usepackage{amsmath}
\usepackage{amssymb}
\usepackage{multirow}
\usepackage{pifont}
\usepackage{subcaption}
\usepackage{xcolor,colortbl}

\title{RECLIP: Resource-efficient CLIP by Training with Small Images}

\author{
\parbox{\linewidth}{\centering
Runze Li\footnotemark[1]\hspace{.4cm}
Dahun Kim \footnotemark[2]\hspace{.4cm}
Bir Bhanu\footnotemark[1]\hspace{.4cm}
Weicheng Kuo \footnotemark[2]\\
}
\parbox{\linewidth}{\centering \vspace{0.2cm}
UC Riverside\footnotemark[1]\hspace{.4cm} Google Deepmind\footnotemark[2]
}
\\
}

\def\month{08}  
\def\year{2023} 
\def\openreview{\url{https://openreview.net/forum?id=Ufc5cWhHko}} 

\usepackage{xspace}
\newcommand*{\eg}{e.g.\@\xspace}
\newcommand*{\ie}{i.e.\@\xspace}
\newcommand{\ours}{RECLIP\xspace}

\newcommand{\revise}[1]{\textcolor{black}{#1}}
\newcommand{\tabincell}[2]{\begin{tabular}{@{}#1@{}}#2\end{tabular}}

\newcommand{\tablestyle}[2]{\setlength{\tabcolsep}{#1}\renewcommand{\arraystretch}{#2}\centering\footnotesize}
\newlength\savewidth\newcommand\shline{\noalign{\global\savewidth\arrayrulewidth
  \global\arrayrulewidth 1pt}\hline\noalign{\global\arrayrulewidth\savewidth}}
\newcommand{\white}[1]{\color[HTML]{FFFFFF}{{#1}}}

\newcommand{\gray}[1]{\textcolor{gray}{{#1}}}

\definecolor{Gray}{gray}{0.85}

\newcommand{\cmark}{\ding{51}}%
\newcommand{\xmark}{\ding{55}}%
\renewcommand{\paragraph}[1]{\vspace{1mm}\noindent\textbf{#1}}
\newcommand\columncolor[1]{\cellcolor{Gray}{#1}}

\begin{document}

\maketitle

\begin{abstract}
We present \ours (Resource-efficient CLIP), a simple method that minimizes computational resource footprint for CLIP (Contrastive Language Image Pretraining). Inspired by the notion of coarse-to-fine in computer vision, we leverage small images to learn from large-scale language supervision efficiently, and finetune the model with high-resolution data in the end. Since the complexity of the vision transformer heavily depends on input image size, our approach significantly reduces the training resource requirements both in theory and in practice. Using the same batch size and training epoch, \ours achieves highly competitive zero-shot classification and image-text retrieval accuracy with 6 to 8$\times$ less computational resources and 7 to 9$\times$ fewer FLOPs than the baseline. Compared to the state-of-the-art contrastive learning methods, \ours demonstrates 5 to 59$\times$ training resource savings while maintaining highly competitive zero-shot classification and retrieval performance. \revise{Finally, \ours matches the state of the art in transfer learning to open-vocabulary detection tasks, achieving 32 AP$r$ on LVIS}. We hope this work will pave the path for the broader research community to explore language supervised pretraining in resource-friendly settings.


\end{abstract}

\section{Introduction}
Representation learning is a foundational problem in computer vision and machine intelligence. Effective image representation can benefit a myriad of downstream tasks, including but not limited to image classification, object detection, semantic segmentation, and 3D scene understanding. In the past decade, the community has witnessed the rise of supervised learning~\citep{imagenet_cvpr09,Sun_2017_ICCV}, then self-supervised learning~\citep{chen20icml,He_2020_CVPR,bao2022beit}, and most recently language-supervised learning~\citep{radford2021clip,align,yu2022coca}. Language-supervised representation gains much traction for its exceptional versatility. It exhibits outstanding performance in zero-shot classification~\citep{radford2021clip}, linear probing~\citep{radford2021clip,yu2022coca}, few-shot learning~\citep{zhou2022coop}, full finetuning~\citep{dong2022clip}, and finds great applications in text-guided image generation~\citep{pmlr-v139-ramesh21a}. Much like the role of supervised pretraining~\citep{imagenet_cvpr09} before, language-supervised pretraining has emerged as a simple yet powerful methodology for representation learning today.

Traditional supervised learning uses a predetermined set of labels, and is effective across a wide range of data and computational resources. In contrast, natural language offers richer learning signals such as object categories or instances, named-entities, descriptions, actions, and their relations at multiple levels of granularity. Unfortunately, this rich supervision also leads to a higher level of noise in the data, where many image-text pairs have only loose connections. To address this noise, data and computational scaling have proven to be highly effective and necessary. For example, training CLIP models require $\sim$3k V100-GPU-days, and likewise CoCa requires $\sim$23k TPU-v4-core-days. Apart from the lengthy training time, the large batch requirement of contrastive learning recipes also demand substantial amount of device memory at all times. These factors limit the research of language supervised learning to institutions with high-end infrastructure, and hinder the exploration by the broader community. 

Thus, improving efficiency of contrastive training has drawn substantial research interest. For example,~\citet{Zhai_2022_CVPR} precomputes the image features by a pretrained classification model to reduce the training cost.~\citet{zhai2023sigmoid} utilizes sigmoid loss to avoid the use of all-gather operation and improves learning with a smaller batch size. Moreover,~\citet{yao2021filip} leverages masked images to speed up contrastive learning. The community have also explored smaller batch sizes~\citep{dong2022maskclip} or curated academic datasets~\citep{li2022supervision,lei2022loopitr} for contrastive learning. However, it is not clear how well the findings in smaller batch and data size settings generalize to larger batch and data size.

\begin{figure*}[t]
    \centering
    \includegraphics[width=0.92\textwidth]{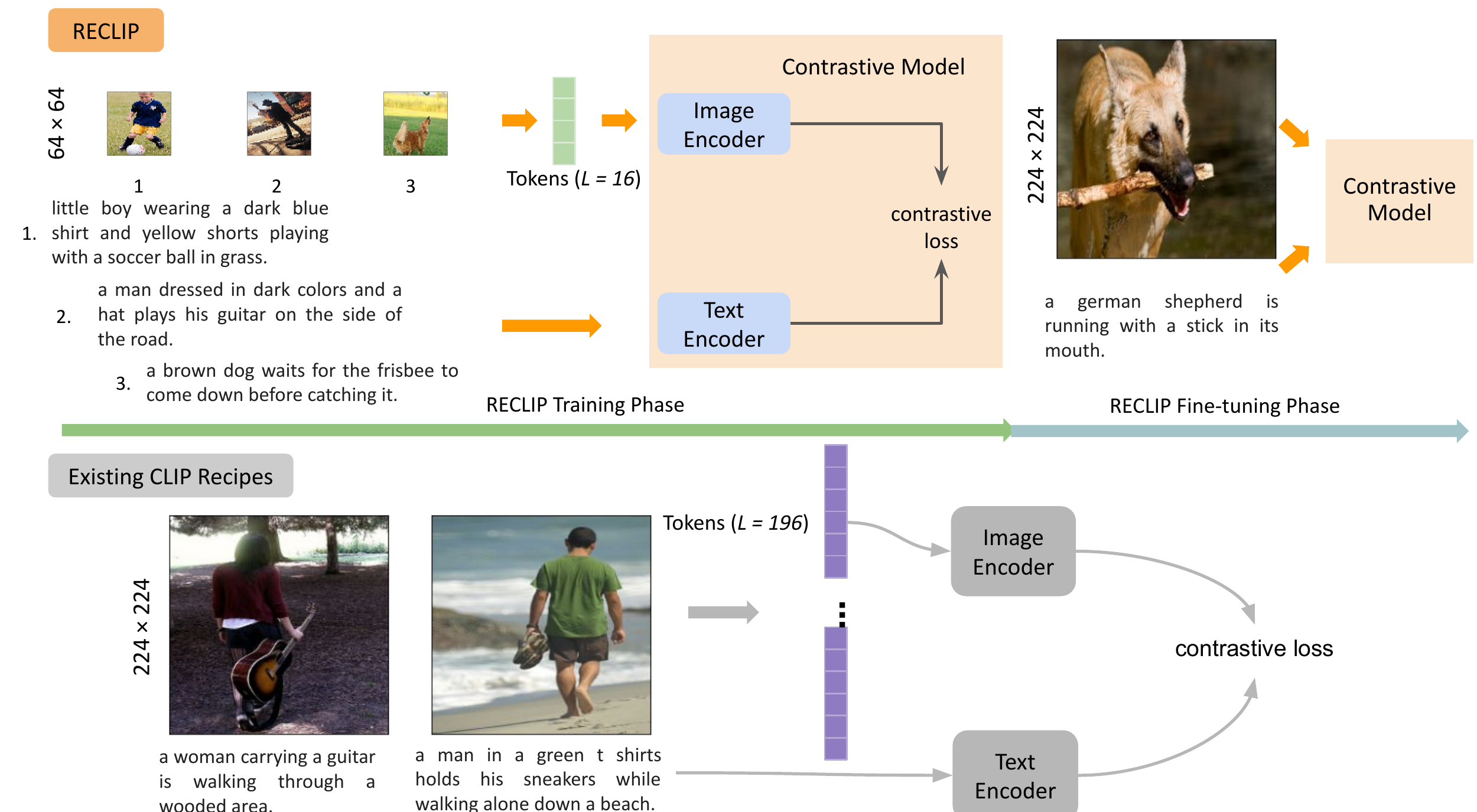}
    \caption{Top: Resource-efficient CLIP (\ours) training pipeline. Bottom: existing CLIP training methods. \ours leverages small images for the main training phase which significantly reduces computational resource requirements through much shorter image sequence length.
    }
    \vspace{-4mm}
    \label{fig:pipeline}
\end{figure*}

We present \ours (Resource-efficient CLIP), a simple method designed to make CLIP more affordable and reproducible for the community (see Fig.~\ref{fig:pipeline}). Consider images 1-3 in the top left of Fig.~\ref{fig:pipeline}. Humans can effortlessly match the images with the corresponding texts below them, \eg ``a boy is playing a soccer ball in grass'' matching image 1. Although the images are only of size $64\times64$, they contain adequate amount of visual information for pairing with texts. Our main insight is to train on small images during the main training phase, and finetune the model with high-resolution images for a short schedule in the end. Intuitively speaking, our approach re-introduces the idea of ``coarse-to-fine'' from classical computer vision to contrastive learning, whereby pretraining incorporates high-level information from small images and finetuning enables the model to refocus its attention on the important details. There is no need for multi-view supervisions~\citep{li2022supervision,yao2021filip}, feature distillation~\citep{lei2022loopitr}, other contrastive losses~\citep{zhai2023sigmoid}, pretrained classifiers~\citep{Zhai_2022_CVPR}, or image masking~\citep{li2022scaling}. Surprisingly, \ours achieves highly competitive zero-shot classification and retrieval performance using $64\times64$ images, which significantly reduces computational resource usage. We attribute this to the complexity of image tower being quartic with respect to the image size (see Eqn.~\ref{eqn:complexity}).

In addition, \ours demonstrates the efficiency and effectiveness of using short sequence length for image language representation learning. Existing \revise{image}-text pretraining methods typically use long sequence lengths, \eg 441~\citep{radford2021clip} or 784~\citep{yu2022coca} to achieve strong downstream zero-shot transfers. Long sequence image encoding has been validated to benefit image classification~\citep{beyer2022better} and object detection~\citep{chen2022simple} with vision transformers. \citet{hu2022exploring} find the sequence length is a key factor for masked image representation learning. Different from these methods that advocate for long sequence length, \ours demonstrates that using only \textbf{16} tokens for the image encoding is sufficient for the main training phase, and can achieve highly competitive zero-shot transfer capabilities via a short high-resolution finetuning schedule. Interestingly, our image sequence length is 4 to 5$\times$ shorter than the \textit{text} sequence lengths of popular recipes \eg 76~\citep{radford2021clip} or 64~\citep{yu2022coca}.

\begin{figure*}[t]
    \centering
    \includegraphics[width=0.93\textwidth]{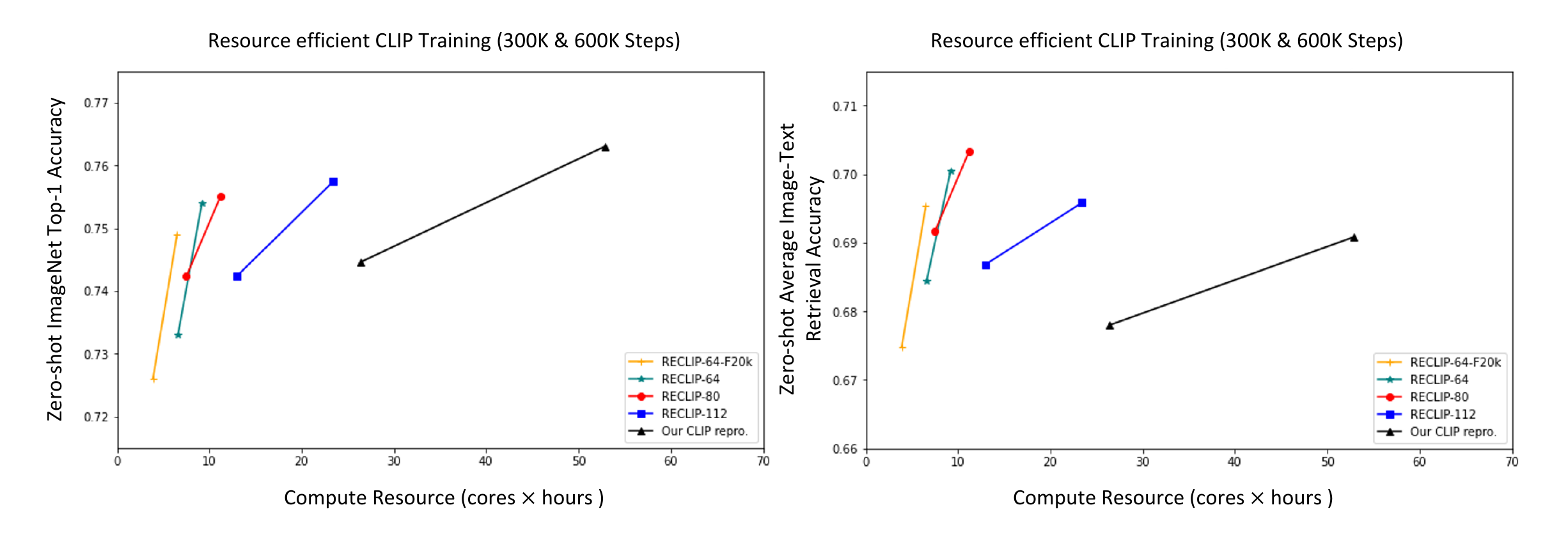}
    \vspace{-2mm}
    \caption{Zero-shot accuracy \textit{vs.} compute resource in cores$\times$hours trade-off. \ours-X: \ours training for 300k and 600k steps with image size $X$ where $X=64, 80, 112$. \ours-64-F20k: \ours-64 finetuned for 20k steps. Our CLIP repro.: our reproduction of CLIP~\citep{radford2021clip}. Zero-shot image-text retrieval results are averaged from image-to-text and text-to-image Recall@1 on two benchmark datasets, Flickr30K~\citep{plummer2015flickr30k} and MSCOCO~\citep{chen2015microsoft}. \ours consumes significantly less compute resource and is more accurate on zero-shot image-text retrieval and highly competitive classification results on ImageNet-1K validation set. 
    }
    \label{fig:teaser_results}
    \vspace{-2mm}
\end{figure*}

In Fig.~\ref{fig:teaser_results}, we present zero-shot classification and retrieval  performance, and resource costs in cores$\times$hours of training \ours models and the baseline model for short and long schedules. Experiments show that, using the same batch size and training steps, \ours reduces the computation resources by 6 to 8$\times$ and largely preserves the classification and retrieval accuracy. When comparing to state-of-the-art (SOTA) methods, \ours significantly saves resource usage by 5 to 59$\times$ and shows highly competitive zero-shot classification and retrieval accuracy. \revise{Apart from image-level tasks, we explore transfer learning of \ours to open-vocabulary detection tasks~\citep{gu2022openvocabulary}, which typically requires high-resolution images for small object recognition. Surprisingly, \ours achieves 32 AP$_r$, matching the state of the art performance of RO-ViT~\citep{Kim_2023_CVPR} on LVIS benchmark. This demonstrates the potential of \ours for region and pixel-level tasks beyond image-level understanding.} In summary, our contributions are:
\begin{itemize}
    \item We present a new language image pretraining methodology, Resource-efficient CLIP (\ours) to minimize computational resource requirements.
    \item We leverage small images for the main contrastive learning phase to enable the model to be trained with language supervisions fast and then finetune the model on high-resolution data with a short schedule in the end.
    \item \ours significantly saves compute resource, reduces FLOPs and achieves highly competitive performance on both zero-shot classification and image-text retrieval benchmarks.
    \item \revise{\ours matches the state of the art in open-vocabulary detection with much less training resources.}
\end{itemize}

We believe \ours could enable the broader research community to explore and understand language supervised pretraining in a more resource friendly setting.

\section{Related Work}

\subsection{Learning with Low-Resolution Images}
Deep learning techniques have been utilized on a wide-variety of computer vision tasks, \eg visual recognition~\citep{He_2016_CVPR,dosovitskiy2021an}, video analysis~\citep{Tran_2018_CVPR}, images generations~\citep{pmlr-v139-ramesh21a}, etc. Most of existing work follow the standard training and testing paradigms to exploit very deep models by using images with the fixed resolution, \eg $224\times 224$. This setting has been one of fundamental standards for various computer vision tasks. However, an increasing number of studies have been conducted to investigate to train deep learning models with low-resolution data. \citet{NEURIPS2019_d03a857a} have observed significant discrepancy on image sizes caused by augmentation methods during the train and test period, and further validated the effectiveness of using lower resolution images for training than testing. Driven by the needs for specific tasks, \eg face recognition, surveillance images analysis, etc., \citet{Singh_2019_ICCV,9453131} and \citet{9681188} study learning with low resolution images and generally focus on using high resolution images as auxiliary data to help to train models with low resolution data, which causes difficulties to generalize on broader visual recognition tasks. For video understanding, \citet{Wu_2020_CVPR} propose to use variable mini-batch shapes with different spatial-temporal resolutions for training deep video models and obtain optimal performance and time trade-offs. With recent advances of vision transformers~\citep{dosovitskiy2021an,he2022masked}, \citet{fastmim} speedup image pretraining by using masked image modelling with low resolution data. \citet{Liu_2022_CVPR} introduce a a log-spaced continuous position bias for pretraining vision models by using smaller images and transfer to high-resolution localization tasks.

\subsection{Language-supervised Learning}
Due to the natural co-occurrence of image and language data on the web, language-supervised learning has become a highly effective and scalable representation learning methodology. Researchers have explored a variety of paired image-text data such as image tags~\citep{chen2015webly,divvala2014learning,joulin2016learning}, captions~\citep{desai2021virtex,sariyildiz2020learning,wang2009learning,ccdataset}, alt-texts~\citep{align,schuhmann2021laion}, image search queries~\citep{radford2021clip}, page title~\citep{chen2022pali}, or a combination of these sources~\citep{chen2022pali}. From a modeling perspective, contrastive learning is particularly suitable for recognition and retrieval tasks, because of its simplicity and versatility. However, the high requirements of computational resources have limited the research from the broader community.

To fully leverage capabilities of vision and language pretraining, large batch size (\eg 16k~\citep{align}, 32k~\citep{radford2021clip,yao2021filip}, or 64k~\citep{yu2022coca}) and web image text data have been adopted widely. This requires a large amount of computational resources which many academic institutions and industry labs cannot afford. To address such limitation,~\citet{Zhai_2022_CVPR} proposes to precompute the image features with frozen classifier backbone, while~\citet{zhai2023sigmoid} proposes sigmoid loss which better supports small batch training. In addition, masked image learning~\citep{yao2021filip}, multi-views data augmentations~\citep{li2022supervision,yao2021filip}, knowledge distillations~\citep{lei2022loopitr} and masked self-distillation~\citet{dong2022maskclip} have been proposed. Since many of these methods are trained and evaluated on smaller scale/data, it is unclear how well they may scale up to larger batch and data. For example, ~\citet{weers2023self} shows that the advantage of contrastive learning approaches on smaller scales may not always hold at larger scales.
In contrast, we propose a simple and novel recipe for language image pretraining, which (1) significantly reduces computation resource requirements and (2) works well on a large-scale web dataset~\cite{chen2022pali} with minimal change to the established CLIP recipe~\citep{radford2021clip}.


\section{Method}
\subsection{Preliminaries}
\paragraph{Contrastive Language Image Pretraining.} \quad Following existing works~\citep{radford2021clip,yu2022coca}, we utilize a transformer-based contrastive model which consists of an image encoder and a text encoder. The image and text encoders are trained to output image-level representation and sentence-level representations respectively. The image embeddings $\{{p}\}$ and text embeddings $\{{q}\}$ are obtained by global average pooling at the last layers of image and text encoders. The cosine similarity of the embeddings in batch $B$, scaled by a learnable temperature $\tau$ are the input to the InfoNCE loss~\citep{oord2018representation,radford2021clip}. The image and text contrastive loss is obtained by $L_{con} = (L_{\text{I2T}} + L_{\text{T2I}}) / 2$, with:
\begin{equation}\label{eqn:contrastive}
L_{\text{I2T}} = -{1 \over {B}} \sum_{i=1}^{B} \log({\text{exp}(p_{i}q_{i} / \tau) \over { \sum_{j=1}^{B} \text{exp}(p_{i} q_{j} / \tau)  }}).
\end{equation}
\begin{equation}\label{eqn:contrastive}
L_{\text{T2I}} = -{1 \over {B}} \sum_{i=1}^{B} \log({\text{exp}(q_{i}p_{i} / \tau) \over { \sum_{j=1}^{B} \text{exp}(q_{i} p_{j} / \tau)  }}).
\end{equation}
where $i,j$ are indexes within the batch. This loss is optimized to learn both the image and language representation in the dual-encoder model.

\subsection{Resource-efficient CLIP}
At a high-level, our method utilizes small images to reduce computation and leverage a brief finetuning stage at the end of training to adapt for high-resolution inference. Intuitively, the use of smaller images presents a trade-off between how much detail we encode per example and how many samples we process per unit of computation resource. 
Fig.~\ref{fig:pipeline} shows the \ours training pipeline on the top. There are two phases: low-resolution main training, and high-resolution finetuning. In the first phase, we leverage small images which contain sufficient visual concepts with paired texts as the input to the image and text encoders. By using an image size of $64$ and a text length of 16, \ours processes the training data significantly faster than existing methods. In the second phase, we finetune the model for a short cycle on high-resolution data to provide valuable image details, which largely enhances the representation quality of the model. Below we delve deeper into specific aspects of \ours design.

\paragraph{Structure preservation by learning from small images.}\quad 
In Fig.~\ref{fig:pipeline}, we observe that small images can preserve visual structure and contain sufficient concepts well. For instance, human can easily tell the object, ``a dog'', in the third image and associate the image with the text of ``a brown dog waits for ...'', and this is a fundamental principle for our \ours to leverage small images for the main language-supervised pretraining. Because down-sampling is a structure-preserving operation \ie global appearance remains similar, we are able to reduce the token length aggressively without compromising the performance of the model. This is different from other techniques to reduce the sequence length (\eg random masking) where the global appearance may change significantly with reduced sequence lengths. Additional visualization presents a comparison between various image resolutions and sheds light on how small images effectively preserve visual appearance (see Fig.~\ref{fig:vis}).

\paragraph{Training complexity with small images.}\quad 
The computation cost of contrastive learning mostly depends on the cost of processing images~\citep{radford2021clip,li2022scaling,yu2022coca}, partly because the image encoder is typically heavier than the text encoder, partly because the image token length tends to be greater than that of text tokens. Below we provide theoretical analysis to understand the efficiency of using small images.

Let the number of tokens from the image encoder be:
\begin{equation}
\label{eqn:image_encoder}
  N = hw / p^2  
\end{equation}, where $h$/$w$ are height/widths of the image, and $p$ is the patch size. If we replace $h$ by $H/r$ and $w$ by $W/r$, where $H/W$ are the original image height and widths, and $r$ is the down-sampling factor. The computation complexity $C$ of the image encoder of a batch is given by
:
\begin{equation}
\label{eqn:complexity}
   C = O(BN^2) = O(\frac{BH^2W^2}{p^2r^4})
\end{equation}
, where $B$ is the batch size. 
When $B, H, W, p$ are held constant, we have:
\begin{equation}
\label{eqn:complexity_downsample}
   C = O(\frac{1}{r^4})
\end{equation}

This shows that reducing the image size is very effective in reducing computation complexity to the inverse power of up to 4. Since image encoder is the computation bottleneck in existing CLIP recipes~\citep{radford2021clip,li2022scaling,Zhai_2022_CVPR,yu2022coca}, \ours reduces the image sequence length to \textbf{16} by using an image size of $64$, which makes our image token length the same as our own text token length, and much shorter than those of aforementioned methods.

\revise{
The above complexity analysis C is calculated based on the core operations self-attention layers in transformers. However, empirically the complexity of a transformer may not be dominated by the self-attention layers, the fully connected layers also play an important role. GPT-3~\citep{NEURIPS2020_1457c0d6} paper have provided computation analysis of their language models, where the computation cost is estimated as $O(N)$, linear with the sequence length. Thus, we discussed the lower-bound of the complexity $C_{lb}$ of RECLIP in~\eqref{eqn:complexity2}. Using the notation of ~\eqref{eqn:complexity} and ~\eqref{eqn:complexity_downsample}, we have
\begin{equation}
\label{eqn:complexity2}
C_{lb} = O(BN) = O(\frac{BHW}{pr^2}),
\end{equation}
During the training, the B, H, W, p are normally constant, so~\eqref{eqn:complexity2} can be simplified as:
\begin{equation}
\label{eqn:complexity_downsample2}
C = O(\frac{1}{r^2}).
\end{equation}
Compared to ~\eqref{eqn:complexity_downsample}, we observe that the computation savings in practice may be somewhere between $O(\frac{1}{r^2})$ and $O(\frac{1}{r^4})$. This analysis shows that changing $r$ is very effective regardless of the compute estimation techniques.}


\paragraph{Constant batch size.} \quad Batch size is a critical factor in contrastive learning~\citep{radford2021clip,basic,li2022scaling,chen2022simple} and larger batch has consistently yielded improvement. In Equation~\ref{eqn:complexity}, the complexity changes linearly with batch size $B$. Observing that reduced batch size tends to hurt representation quality, we keep the batch size constant to save both computation and memory use by reducing image size only.

\paragraph{High-resolution finetuning.} \quad We perform high resolution finetuning after the main low-resolution training. Intuitively speaking, the model has acquired a high-level understanding of the images and texts through the main training phase. We improve its representation further by providing more detailed visual information through a short high-resolution finetuning process. \revise{ The images used for high-resolution training are the same as those for the low-res training, except that we remove the downsampling to preserve the rich visual details.}

Care is taken to initialize the positional embeddings from low-res pretraining to high-res finetuning. We up-sample the positional embedding weights from the low dimension (\eg 4x4 for $h=w=64$) to the dimension of high-resolution positional embeddings (\eg 14x14 for $h=w=224$) for a given patch size $p=16$. Compared to up-sampling the low-res positional embeddings without increasing the amount of weights, we found this weight up-sampling beneficial because the positional embeddings have higher capacity to adapt with more detailed spatial representation. \revise{We use trainable positional embedding throughout the paper following existing works~\citep{radford2021clip,dosovitskiy2021an,yu2022coca}.}

\paragraph{Network architecture.} \quad We use the ViT-Large backbone as image encoder by default unless noted otherwise. \revise{The ViT-Large is a vision transformer which consists of 24 multi-head self-attention layers with 16 heads and the width dimension of 1024.} The patch size is fixed at $16$ following common practice. Although we focus on ViT architecture in this study, \ours involves only changing the input size, and can potentially support other network architectures as well~\citep{NIPS2017_3f5ee243,dosovitskiy2021an,He_2016_CVPR,Liu_2021_ICCV,NEURIPS2021_cba0a4ee}. Our text encoder follows the same transformer design as previous works~\citep{radford2021clip,yu2022coca}. \revise{The text encoder consists of 12 multi-head self-attention layers with 12 heads and the width dimension of 1024.} 

\paragraph{Implementation details.} \quad We use a starting learning rate of 0.001, and train for 250k and 550k steps with linear LR decay using an Adafactor optimizer. We set weight decay to 0.01 and batch size to 16384. \revise{The batch size is chosen to be a multiple of 1024 and the model feature dimension (e.g. 4096) a multiple of 128, so that TPU padding would not occur on the sequence dimension.} A short LR warmup of 2500 steps is used. Our high-resolution finetuning schedule starts with a learning rate of \revise{$10^{-4}$} with 5000 steps LR warmup, and decays linearly over a total schedule of 20k or 50k iterations. We use an image size of 224 or 448 for finetuning. We use the English subset of the WebLI dataset~\citep{chen2022pali} for training. Our training is run on TPU-v3 infrastructure. \revise{Compared to general-purpose GPU devices, TPUs are specifically designed for large matrix operations commonly used in neural networks. Each TPU v3 device has 16GB high-bandwidth memory per core, which is comparable to that of a V100 and suitable for synchronous large-scale training.} For zero-shot image classification, we use the same text prompts as~\citet{radford2021clip}.


\section{Experiments}

\subsection{Main Results}



\paragraph{Zero-shot image-text retrieval and image classification.}\quad
Following existing works~\citep{radford2021clip,li2022scaling,yu2022coca}, we evaluate \ours on zero-shot image and text retrieval on Flickr30K~\citep{plummer2015flickr30k} and MSCOCO~\citep{chen2015microsoft} test sets, and zero-shot image classification on ImageNet~\citep{imagenet_cvpr09}, ImageNet-A~\citep{hendrycks2021nae}, ImageNet-R~\citep{hendrycks2021many}, ImageNet-V2~\citep{pmlr-v97-recht19a} and ImageNet-Sketch~\citep{wang2019learning} datasets. we take each image and text to the corresponding encoder to obtain embeddings for all image and text pairs. Then we calculate the the cosine similarity scores for the retrieval, and use the aligned image and text embeddings to perform zero-shot image classification by matching images with label names without fine-tuning.

Table~\ref{tab:main_results} presents the results of \ours on this benchmark, where the baseline is our own reproduced version of CLIP. Our baseline model trains on the WebLI dataset with the images of $224\times224$ for 300k and 600k steps. The original CLIP~\citep{radford2021clip} model and trains on their own dataset with the image size of $336\times336$, which is marked in \gray{gray}. \ours uses small images for the main training phase and finetune the model with the images of $224\times224$ for 20k or 50k steps.

For long-schedule training of 600k steps, \ours-64 significantly reduces compute use by $\bf\sim6$ times from 52.8K to \textbf{9.2K} in cores$\times$ hours, which saves $\bf\sim80\%$ compute resource, and it outperforms the baseline model by +3.9 on Flickr and MSCOCO retrieval. \ours-64-F20K, which finetunes the model for only 20k steps with high-resolution images, further reduces the computation use by $\bf\sim8\times$ to \textbf{6.5K} and improves retrieval performance by +1.8. On zero-shot image classification, \ours-64 achieves 75.4 and \ours-64-F20K achieves 74.9 of the top-1 accuracy, which is very competitive with the baseline method. \ours-64 reduces the token length for the image encoding from 196 to \textbf{16} during the main training phase, which is a key factor for resource savings.  Overall, \ours-64 shows attractive trade-offs between the resource use and zero-shot retrieval and image classification performance. 

We also train \ours with the image size of $80\times80$. Comparing to the baseline method which consumes 52.8K in cores$\times$hours, our \ours-80 remarkably reduces resource usage by $\bf\sim5$ times to \textbf{11.2K}. \ours-80 improves retrieval results by +5.0 on Flickr30K and MSCOCO test sets, and achieves highly competitive zero-shot image classification performance of 75.8. Specifically, taking INet-A as an example, \ours-80 outperforms the baseline method for both 300k and 600k training steps. For short training schedule with 300 steps, \ours-80 requires only \textbf{7.5K} in cores$\times$hours which is $\bf\sim4\times$ less than the baseline model.

\begin{table*}[t]
    \caption{Zero-shot image-text retrieval, image classification results. CLIP*: The original CLIP model~\citep{radford2021clip} is marked in \gray{gray}. The resource use is converted to TPU-v3 core-hours per~\cite{li2022scaling}. CLIP, our repro.: our reproduced CLIP. \ours-{$X$}: \ours trained with image size $X$ where $X=64, 80, 112$. RECLIP-64-F20K: \ours-64 finetuned for a shorter schedule of 20k steps. \revise{Best results are \textbf{bolded}.}}
    \vspace{-2mm}
    \centering
    \small
    \resizebox{\textwidth}{!}{
    \begin{tabular}{l|c|c|cc|cc|cccccc}
    \multirow{3}{*}{Method} & \multirow{3}{*}{\tabincell{c}{Training \\ steps}} & \multirow{3}{*}{\tabincell{c}{Cores \\$\times$hours}} & \multicolumn{4}{c|}{Zero-shot Retrieval} & \multicolumn{5}{c}{Zero-shot INet Classification}
    \\
    ~ & ~ & ~ & \multicolumn{2}{c|}{Flickr30K (1K test set)} & \multicolumn{2}{c|}{MSCOCO (5K test set)} & \multirow{2}{*}{INet} & \multirow{2}{*}{INet-A} & \multirow{2}{*}{INet-R} & \multirow{2}{*}{INet-V2} & \multirow{2}{*}{INet-Sketch}\\
    ~ & ~ & ~ & I2T R@1 & T2I R@1 & I2T R@1 & T2I R@1 & ~ & ~ & ~ & ~ & ~  \\
    \shline
    \gray{CLIP*~\citep{radford2021clip}} & - & \gray{120.0K} & \gray{88.0} & \gray{68.7} & \gray{58.4} & \gray{37.8} & \gray{76.2} & \gray{77.2} & \gray{88.9} & \gray{70.1} & \gray{60.2} \\
    \hline 
    CLIP, our repro. & 300k & 26.4K & 89.3 & 75.4 & 61.3 & 45.1 & \bf74.5 & 54.4 & \bf88.9 & \bf67.7 & \bf64.5 \\
    \hline
    \bf{\ours-112} & 300k & 13.1K & 90.0 & 76.6 & \bf63.1 & 45.0 & 74.2 & 55.4 & 87.8 & 67.2 & 63.2 \\
    \bf{\ours-80} & 300k & 7.5K & \bf91.0 & \bf77.1 & 62.8 & \bf45.7 & 74.3 & \bf56.7 & 87.8 & 67.2 & 62.9 \\
    \bf{\ours-64} & 300k & 6.6K & 89.4 & 77.0 & 62.2 & 45.2 & 73.3 & 53.7 & 86.3 & 66.2 & 61.6 \\
    \bf{\ours-64-F20K} & 270k & \bf3.9K & 88.5 & 76.1 & 60.8 & 44.5 & 72.6 & 51.7 & 85.3 & 65.3 & 60.6 \\
    \shline
    CLIP, our repro. & 600k & 52.8K & 89.3 & 76.9 & 63.3 & 46.8 & \bf76.4 & 60.2 & \bf90.9 & \bf70.1 & \bf66.4 \\
    \hline
    \bf{\ours-112} & 600k & 23.4K & 90.6 & 77.6 & 63.6 & 46.5 & 75.8 & 58.8 & 89.3 & 69.1 & 65.2 \\
    \bf{\ours-80} & 600k & 11.2K & \bf91.3 & \bf78.2 & \bf64.6 & \bf47.2 & 75.8 & 60.3 & 89.0 & 69.2 & 64.6 \\
    \bf{\ours-64} & 600k & 9.2K & 91.0 & 78.1 & 64.2 & 46.9 & 75.4 & \bf60.9 & 88.8 & 68.9 & 64.5 \\
    \bf{\ours-64-F20K} & 570k & \bf6.5K & 91.0 & 77.1 & 63.6 & 46.2 & 74.9 & 58.6 & 88.2 & 68.4 & 63.5 \\
    \hline
    \end{tabular}
    }
    \label{tab:main_results}
    \vspace{-4mm}
\end{table*}

\vspace{-2mm}
\begin{table}[!h!]
    \caption{Comparisons of GFLOPs between \ours and the baseline model during the \ours training. }
    \vspace{-2mm}
    \centering
    \tablestyle{12pt}{1.2}
    \begin{tabular}{l|c}
    Models & GFLOPs \\
    \hline
    CLIP, our repro. & 71.4 \\
    \bf{\ours-112} & 24.8  \\
    \bf{\ours-80} & 10.1  \\
    \bf{\ours-64} & 7.3  \\
    \hline
    \end{tabular}
    \label{tab:gflops}
    \vspace{-2mm}
\end{table}

\paragraph{GFLOPS.}\quad
We compare GFLOPs of \ours with the baseline method in Table~\ref{tab:gflops}. The baseline method, CLIP, our repro., requires 71.4 GFLOPs. Our \ours-80 reduces GFLOPs by $\bf\sim7\times$ to \textbf{10.1} and \textbf{\ours-64} further reduces GFLOPs by $\bf\sim10\times$ by using even smaller images.


\subsection{System-level Comparison}
We present system-level comparison between \ours and a series of existing methods on Flickr30K and MSCOCO image-text retrieval benchmarks, and ImageNet classification accuracy in Table~\ref{tab:sota}. We train \ours for 600k steps and then finetune for 50k steps with the image size of $448\times448$. For \ours-64-F20K, we finetune for 20k steps.

From Table~\ref{tab:sota}, we observe clear resource savings and highly competitive performance achieved with our simple and efficient training recipes. \ours with small images saves $3\sim59\times$ compute resource in cores$\times$hours. When comparing to the models with the similar scale of the image encoder~\citep{radford2021clip,align,yao2021filip,li2022scaling}, \ours reduces resource use by $\bf5\sim22$ times with competitive zero-shot retrieval and image classification performance. In the comparisons to FLIP~\citep{li2022scaling}, \ours-64-F20K uses $\bf{\sim5\times}$ less resource in cores$\times$hours and outperforms it by +2.0 on Flickr30k and MSCOCO retrieval. Surprisingly, when compared to the CoCa, \ours-64-F20K significantly saves $\bf\sim98\%$ resource use and achieves the best image to text retrieval on Flickr30K test set, giving 92.5 of R@1. \ours-64-F20K gives 75.3, which is very competitive on zero-shot ImageNet classification among purely language supervised approaches. We believe this resource savings mostly come from the use of very short image sequence length \ie, 16, which is very different from existing recipes~\citep{radford2021clip,li2022scaling,yu2022coca,Zhai_2022_CVPR}.

We also observe that \ours-80 uses $3\sim34\times$ less compute resource. When comparing to the CoCa~\citep{yu2022coca}, \ours-80 saves $\bf{\sim97\%}$ resource use and achieves highly competitive retrieval performance. The resource savings of \ours-80 can also be attributed to the largely-reduced sequence length, \ie, \textbf{25} for the image encoding. \ours-80 achieves highly competitive ImageNet top1 accuracy of 76.3, which outperforms CLIP and is on-par with ALIGN. Overall, \ours provides very affordable recipes for large-scale language and image pretraining.

We note that some leading methods~\citep{chen2022pali,basic,yu2022coca} marked in \gray{gray} demonstrate substantially better zero-shot classification because of larger image encoder capacity and the use of JFT~\citep{Sun_2017_ICCV} dataset. JFT is a human-annotated classification dataset which is cleaner than most web crawled image-text datasets~\citep{radford2021clip,schuhmann2021laion,align} and most advantageous for zero-shot classification, so we list the JFT-trained entries there for reference only. 

\begin{table*}[t]
    \caption{Comparisons of zero-shot image-text retrieval and ImageNet classification top-1 accuracy on Flickr30K, MSCOCO and ImageNet. Models that use the fully-supervised dataset~\citep{Sun_2017_ICCV} and much larger are marked in \gray{gray}. $\dagger$: We refer to ~\citep{li2022scaling} to convert GPU cost to TPU usage in CLIP~\citep{radford2021clip}, FILIP~\citep{yao2021filip}. Cores$\times$hours results are reported on TPU-v3 infrastructure. \revise{Best results are \textbf{bolded}.} }
    \vspace{-2mm}
    \centering
    \small
    \resizebox{\textwidth}{!}{
    \begin{tabular}{l|c|c|c|cccc|cccc}
    ~ & Image & Cores & ~ & \multicolumn{4}{c|}{Flickr30K (1K test set)} & \multicolumn{4}{c}{MSCOCO (5K test set)} \\
    ~ & Encoder & $\times$ & ImageNet &  \multicolumn{2}{c}{\underline{{\white{---}}image-to-text{\white{---}}}} & \multicolumn{2}{c|}{\underline{{\white{---}}text-to-image{\white{---}}}} & \multicolumn{2}{c}{\underline{{\white{---}}image-to-text{\white{---}}}} & \multicolumn{2}{c}{\underline{{\white{---}}text-to-image{\white{---}}}} \\
    Method & Size & Hours & Top-1 & R@1 & R@5 & R@1 & R@5 & R@1 & R@5 & R@1 & R@5 \\
    \shline
    \gray{PaLI~\citep{chen2022pali}} & \gray{3.9B} & \gray{598.7K} & \gray{85.4} & - & - & - & - & - & - & - & -  \\
    \gray{BASIC~\citep{basic}} & \gray{2.4B} & \gray{288.1K} & \gray{85.7} & - & - & - & - & - & - & - & -  \\
    \gray{CoCa~\citep{yu2022coca}} & \gray{1B} & \gray{962.1K} & \gray{\bf86.3} & \gray{\bf92.5} & \gray{\bf99.5} & \gray{\bf80.4} & \gray{\bf95.7} & \gray{\bf66.3} & \gray{\bf86.2} & \gray{\bf51.2} & \gray{\bf74.2} \\

    \hline
    CLIP~\citep{radford2021clip} & 302M & 120.0K$\dagger$ & 76.2 & 88.0 & 98.7 & 68.7 & 90.6 & 58.4 & 81.5 & 37.8 & 62.4 \\
    ALIGN~\citep{align} & 408M & 355.0K & 76.4 & 88.6 & 98.7 & 75.7 & 93.8 & 58.6 & 83.0 & 45.6 & 69.8  \\
    FILIP~\citep{yao2021filip} & 302M & 180.0K$\dagger$ & \bf{78.3} & 89.8 & \bf{99.2} & 75.0 & 93.4 & 61.3 & 84.3 & 45.9 & 70.6 \\
    FLIP~\citep{li2022scaling} & 303M & 81.9K & 75.8 & 91.7 & - & 78.2 & - & 63.8 & - & 47.3 & -  \\
    \hline

    \bf{\ours-80 (ours)}  & 303M & 28.7K & 76.3 & 91.4 & 99.1 & \bf79.2 & 94.7 & \bf64.9 & \bf85.2 & \bf48.2 & \bf72.6  \\
    \bf{\ours-64-F20K (ours)}  & 303M & \bf16.4K & 75.3 & \bf92.5 & 99.1 & 78.7 & \bf94.9 & 64.5 & \bf85.2 & 47.3 & 71.9  \\
    \hline
    \end{tabular}
    }
    \label{tab:sota}
    \vspace{-2mm}
\end{table*}



\begin{table}[t]
    \caption{\revise{LVIS open-vocabulary object detection. RECLIP maintains the same open-vocabulary detection (AP$_r$) and standard detection (AP) as the state of the art RO-ViT despite using much less training resources.}}
    \centering
    \tablestyle{12pt}{1.2}
    \begin{tabular}{l|l|l|c|c}
    ViT based method & Pretrained model & Detector backbone & AP$_r$ & AP  \\
    \hline
    RO-ViT~\citep{Kim_2023_CVPR} & ViT-L/16 & ViT-L/16 & \bf 32.1 & 34.0 \\
    \bf RECLIP-RO-ViT (Ours) & ViT-L/16 & ViT-L/16 & 32.0 & \bf 34.7  \\
    \hline
    \end{tabular}
    \label{tab:open_vocab}
    \vspace{-2mm}
\end{table}

\subsection{\revise{Open Vocabulary Detection}}

\revise{We conduct evaluation on the LVIS dataset~\citep{Gupta_2019_CVPR} by using RECLIP for open vocabulary detection. We take a recent SOTA approach RO-ViT~\citep{Kim_2023_CVPR} as the baseline and apply RECLIP-80 to pre-train the model (RECLIP-RO-ViT). We train only on the LVIS base categories (frequent \& common) and test on both the base and novel (rare) categories following the protocol of ViLD~\citep{gu2022openvocabulary}. The results are in the Table~\ref{tab:open_vocab}. RECLIP-RO-ViT achieves 32.0 Mask APr (AP on rare categories)~\citep{Gupta_2019_CVPR}, matching the state of the art performance of RO-ViT (32.1). This is surprisingly encouraging because detection task typically requires much higher resolution e.g. 1024 than classification task to recognize the small objects, which can be especially challenging for RECLIP due to the low-res information loss. In addition, RECLIP-RO-ViT outperforms RO-ViT by 0.7 on all-category AP, showing that its representation is also suitable for standard detection on the base categories. These detection results suggest that RECLIP representation is versatile and suitable for a broader range of object and pixel-level tasks.} 

\subsection{Ablations}
In this section, we ablate the design of \ours training and evaluate on the zero-shot retrieval and classification accuracy.

\paragraph{The importance of high-resolution finetuning.}\quad
Table~\ref{tab:ablation_highres_finetuning} shows the importance of finetuning \ours with high-resolution data after the main training phase. We compare the retrieval and classification accuracy by using the model trained with and without high-resolution finetuning on an image size of $224$ for 50k steps. We observe that high-resolution finetuning significantly improves the performance for zero-shot retrieval and classification. In particular, training \ours by using the smallest images, \eg $64\times64$, high-resolution finetuning offers the most notable benefits. This is also aligned with the results in Table~\ref{tab:sota} where \ours models trained with small images, \eg $64\times64$ or $80\times80$, and finetuned with $448\times448$ for a short cycle can achieve comparable performance with SOTA models.

\vspace{-2mm}
\begin{table}[!h]
    \caption{The importance of \ours high-resolution finetuning. We found that high-resolution finetuning significantly improves zero-shot transfer performance. \ours-X: \ours trained with image size $X$. \revise{Best results are \bf{bolded}.}}
    \vspace{-2mm}
    \centering
    \tablestyle{10pt}{1.2}
    \resizebox{\textwidth}{!}{
    \begin{tabular}{c|c|ccccc|ccccc}
    ~ & Total & \multicolumn{5}{c|}{Before high-resolution finetuning} & \multicolumn{5}{c}{After high-resolution finetuning} \\ 
    ~ & Training & INet & \multicolumn{2}{c}{Flickr30K} & \multicolumn{2}{c|}{MSCOCO} & INet & \multicolumn{2}{c}{Flickr30K} & \multicolumn{2}{c}{MSCOCO} \\
    ~ & Steps & Top-1 & I2T & T2I & I2T & T2I & Top-1 & I2T & T2I & I2T & T2I  \\
    \hline
    \ours-112 & 300k & \bf{69.0} & \bf{83.2} & \bf{67.9} & \bf{58.6} & \bf{40.0} & 74.2 (+5.2) & 90.0 (+6.8) & 76.6 (+8.7) & \bf63.1 (+4.7) & 45.0 (+5.0) \\
    \ours-80 & 300k & 66.3 & 80.8 & 65.4 & 54.6 & 37.4 & \bf74.3 (+8.0) & \bf91.0 (+10.2) & \bf77.1 (+11.7) & 62.8 (+8.2) & \bf45.7 (+8.3) \\
    \ours-64 & 300k & 62.8 & 79.6 & 63.6 & 51.4 & 34.5 & 73.3 (+10.5) & 89.4 (+9.8) & 77.0 (+6.4) & 62.2 (+10.8) & 45.2 (+10.7) \\
    \hline
    \ours-112 & 600k & \bf{70.7} & \bf{87.4} & \bf{71.9} & \bf{59.0} & \bf{40.9} & \bf75.8 (+5.1) & 90.6 (+3.2) & 77.6 (+5.7) & 63.6 (+4.6) & 46.5 (+5.5) \\
    \ours-80 & 600k & 67.7 & 82.8 & 68.1 & 55.8 & 39.0 & \bf75.8 (+8.1) & \bf91.3 (+8.3) & \bf78.2 (+10.1) & \bf64.6 (+ 8.8) & \bf47.2 (+8.2) \\
    \ours-64 & 600k & 65.5 & 80.9 & 66.1 & 54.3 & 37.1 & 75.4 (+9.9) & 91.0 (+10.1) & 78.1 (+12.0) & 64.2 (+10.1) & 46.9 (+9.8)   \\
    \hline
    \end{tabular}
    }
    \label{tab:ablation_highres_finetuning}
    \vspace{-2mm}
\end{table}

\vspace{-2mm}
\paragraph{Text length for \ours main training.}\quad
Table~\ref{tab:ablation_text_length} studies the text length for the \ours training. We use the text length of 64 and 16 to train our \ours with an image size of $80$. Somewhat surprisingly, we observe that using a short text length, \ie 16, during the main training phase clearly reduces the resource use and achieve competitive zero-shot retrieval and image classification performance. This training efficiency gains is possible because we use much shorter image sequence lengths than existing recipes~\citep{radford2021clip,yu2022coca}. 

\vspace{-2mm}
\begin{table}[!h]
    \caption{The effect of the text length in \ours main training. We found that using a short image sequence can further save compute resource and achieve promising zero-shot transfer performance. Default \ours settings are in \colorbox{Gray}{dark gray}. Best results are \textbf{bolded}.}
    \vspace{-2mm}
    \centering
    \tablestyle{12pt}{1.2}
    \begin{tabular}{c|c|cc|cc|c}
    Text & Cores & \multicolumn{2}{c|}{Flickr30K} & \multicolumn{2}{c|}{MSCOCO} & INet \\
     Length & $\times$ hours & I2T & T2I & I2T & T2I & Top-1\\
    \hline
     64 & 15.5K & 91.2 & 78.0 & 64.3 & 46.7 & 75.6 \\
     \rowcolor{Gray}
     16 & \textbf{11.2K} & \textbf{91.3} & \textbf{78.2} & \textbf{64.6} & \textbf{47.2} & \textbf{75.8} \\
    \hline
    \end{tabular}
    \label{tab:ablation_text_length}
    \vspace{-2mm}
\end{table}

\paragraph{Small batch size for \ours main training phase.}\quad
Our \ours is designed with principles of using constant batch size but varying image resolutions during the main training phase. Table~\ref{tab:ablation_batch_size} ablates effects of the batch size during the main training phase on zero-shot retrieval and image classification accuracy. We first train the model for 250k steps by using the batch size of 4k or 16k and the image size $112$; then we finetune it for 50k steps by using the batch size of 16k and the image size of $224$.
From Table~\ref{tab:ablation_batch_size} shows that using smaller batch size (4k)  saves compute resource by $69\%$, but the zero-shot retrieval and classification performance drops significantly even with the same high-resolution finetuning phase. Therefore, we conclude that using the same large batch size is important for language image pretraining to ensure competitive zero-shot transfer performance.

\vspace{-2mm}
\begin{table}[!h]
    \caption{The importance of \ours main training with constant batch size. We found that using the same batch size (16k) for \ours main training and finetuning achieves better zero-shot transfer performance. Default \ours settings are in \colorbox{Gray}{dark gray}. Best results are \textbf{bolded}.}
    \vspace{-2mm}
    \centering
    \small
    \tablestyle{12pt}{1.2}
    \begin{tabular}{c|c|cc|cc|c}
    Batch & Cores$\times$ & \multicolumn{2}{c|}{Flickr30K} & \multicolumn{2}{c|}{MSCOCO} & INet \\
    Size & Hours & I2T & T2I & I2T & T2I & Top-1 \\
    \hline
    4k & \textbf{4.2K} & 81.9 & 68.8 & 51.2 & 38.6 & 64.4 \\
    \hline
    \rowcolor{Gray}
    16k & 13.1K & \textbf{90.0} & \textbf{76.6} & \textbf{63.1} & \textbf{45.0} & \textbf{74.2} \\
    \hline
    \end{tabular}
    \label{tab:ablation_batch_size}
    \vspace{-2mm}
\end{table}

\paragraph{Increasing the batch size with small images for \ours}.\quad
In Table~\ref{tab:ablation_mg}, we ablate \ours by varying both the batch size and image size during the main training phase. The multi-grid training paradigm is as below: (1) we equally divide training process into 3 stages with the same steps in each; (2) we train the model for 25k, 50k and 100k steps by using the batch size of 64k, 32k and 16k, and the image size of $112$, $160$ and $224$ in each stage. The idea is to increase the batch size while using low resolution data, and decrease the batch size with high-resolution data. The multi-grid free baseline is trained for 300k steps by using a constant batch size 16k and image size $112$, and finetuned with image size $224$. We observe that \ours without ``MG" is not only simpler, but saves computational resource by $30\%$. In addition, \ours achieves better zero-shot retrieval retrieval performance on Flickr30K and MSCOCO and very similar ImageNet performance.

\vspace{-2mm}
\begin{table}[!h]
    \caption{The effect of multigrid training strategy, where we increase the image size and decrease the batch size simultaneously. We found \ours is simple and effective. Default \ours settings are in \colorbox{Gray}{dark gray}.}
    \vspace{-2mm}
    \centering
    \small
    \tablestyle{12pt}{1.2}
    \begin{tabular}{c|c|cc|cc|c}
    \multirow{2}{*}{MG} & Cores$\times$ & \multicolumn{2}{c|}{Flickr30K} & \multicolumn{2}{c|}{MSCOCO} & INet \\
    ~ & Hours & I2T & T2I & I2T & T2I & Top-1 \\
    \hline
    \cmark & 18.4K & 89.2 & 75.5 & 62.3 & \textbf{45.3} & \textbf{74.5} \\
    \hline
    \rowcolor{Gray}
    \xmark & \textbf{13.1K} & \textbf{90.0} & \textbf{76.6} & \textbf{63.1} & 45.0 & 74.2
\\
    \hline
    \end{tabular}
    \label{tab:ablation_mg}
    \vspace{-2mm}
\end{table}

\paragraph{Multi-stages \ours high-resolution finetuning.}\quad
In Table~\ref{tab:ablation_multi_stage}, we further study \ours with 1 and 2 high-resolution finetuning stages given a model trained with low-resolution data. We study the following two variants. ($112\rightarrow224\rightarrow448$): we train the model for 300k steps with the image size of $112$, finetune it for 40k steps with the image size of $224$, and finetune it for another 40k steps with the image size of $448$. ($112\rightarrow448$): we train the model for 300k steps with the image size of $112$ and finetune it for 50k steps with the image size of $448$. We set 50k steps to keep the computation cost comparable with the first one. We observe that ($112\rightarrow448$) gives very competitive zero-shot retrieval and image classification accuracy. Thus, we use only one high-resolution finetuning stage.

\vspace{-4mm}
\begin{table}[!h]
    \caption{\ours with one-stage or multi-stages high-resolution finetuning. We found that one high-resolution finetuning stage is simple and sufficient. Default \ours settings are in \colorbox{Gray}{dark gray}. The best results are \textbf{bolded}. }
    \vspace{-2mm}
    \centering
    \small
    \tablestyle{12pt}{1.2}
    \begin{tabular}{c|c|cc|cc|c}
    \multirow{2}{*}{Stages} & Core$\times$ & \multicolumn{2}{c|}{Flickr30K} & \multicolumn{2}{c|}{MSCOCO} & INet \\
    ~ & Hours & I2T & T2I & I2T & T2I & Top-1 \\
    \hline
    $112\rightarrow224\rightarrow448$ & 31.1K & 91.0 & 77.7 & 64.1 & \textbf{47.4} & \textbf{76.2} \\
    \hline
    \rowcolor{Gray}
    $112\rightarrow448$ & \textbf{30.8K} &  \textbf{90.7} & \textbf{78.0} & \textbf{64.3} & 47.0 & 76.1 \\
    \hline
    \end{tabular}
    \label{tab:ablation_multi_stage}
    \vspace{-2mm}
\end{table}

\paragraph{Comparisons of image resizing and token masking}\quad \revise{In Table~\ref{tab:ablation_resizing_masking}, we present a comparison between token masking~\citep{li2022scaling} and image resizing training strategy with matching computational budget. The benchmark is zero-shot ImageNet classification. All factors other than masking vs resizing are controlled to be the same. For example, we use the same batch size, data, training recipe, and the same number of iterations for low-resolution (vs masked) pretraining and high-resolution (vs unmasked) finetuning. To match the compute usage betweeen resizing and masking, we set the masking ratios such that the sequence lengths are the same. For example, Mask-112 masks 75\% tokens to match the sequence length of RECLIP-112 (assuming the baseline using full image size 224x224). Table~\ref{tab:ablation_resizing_masking} shows that image resizing has a clear advantage over token masking. RECLIP-112 starts with a gap of +2.9 with Mask-112. As the token masking ratio goes above 75\% (Mask-112), we observe an increasing gap between resizing and masking (+5.4\% for RECLIP-64), showing the clear advantage of resizing in very low-compute settings.}

\begin{table}[!h]
    \caption{\revise{Comparison of resizing vs token masking on zero-shot ImageNet classification. RECLIP-X: RECLIP with image size X. Mask-X: token masking with the same compute budget as the corresponding RECLIP-X. Resizing consistently outperforms masking, and the gap increases with decreasing compute budget. Best results are \textbf{bolded}.}}
    \vspace{-2mm}
    \centering
    \small
    \tablestyle{10pt}{1.2}
    \begin{tabular}{c|c|c}
    X (Image Size) & Mask-X & \columncolor{RECLIP-X} \\
    \hline
    112 & 72.9 & \columncolor{\bf 75.8 (+2.9)}  \\
    80 & 71.3 & \columncolor{\bf 75.8 (+3.5)} \\
    64 & 69.5 & \columncolor{\bf 74.9 (+5.4)} \\ 
    \hline
    \end{tabular}
    \label{tab:ablation_resizing_masking}
    \vspace{-2mm}
\end{table}

\begin{figure*}[t!]
    \centering
    \includegraphics[width=0.85\textwidth]{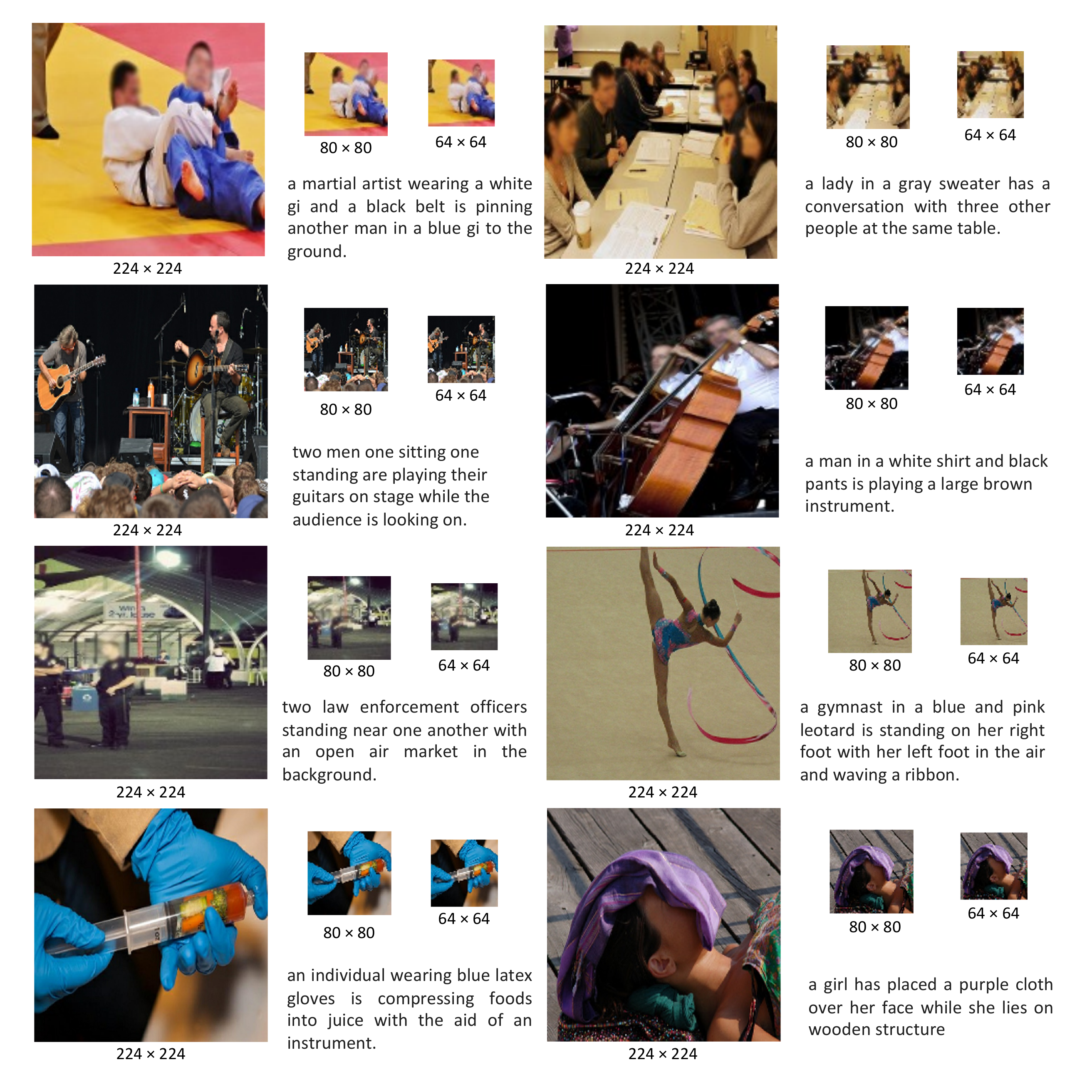}
    \vspace{-3mm}
    \caption{Visualization of image-text pairs and images are in various resolutions. Images are scaled with the same factor of 0.01 for both height and width. Small images contain sufficient visual information for contrastive training.}
    \label{fig:vis}
    \vspace{-4mm}
\end{figure*}

\subsection{Visualization}
\label{sec:visualization}
\paragraph{Visualization of small images.}\quad
In Fig.~\ref{fig:vis}, we visualize images at various resolutions paired with their corresponding texts. We observe that small images generally preserve high-level structures of the original images, and contain sufficient visual information for language supervisions. For example, the martial arts, office meeting, concert, and gymnastics scenes are clearly recognizable down to $64\times64$ resolution. This supports the key insight of our \ours training design that leverages small images for the main training phase to save computation.

\begin{figure*}[t!]
    \centering
    \includegraphics[width=0.85\textwidth]{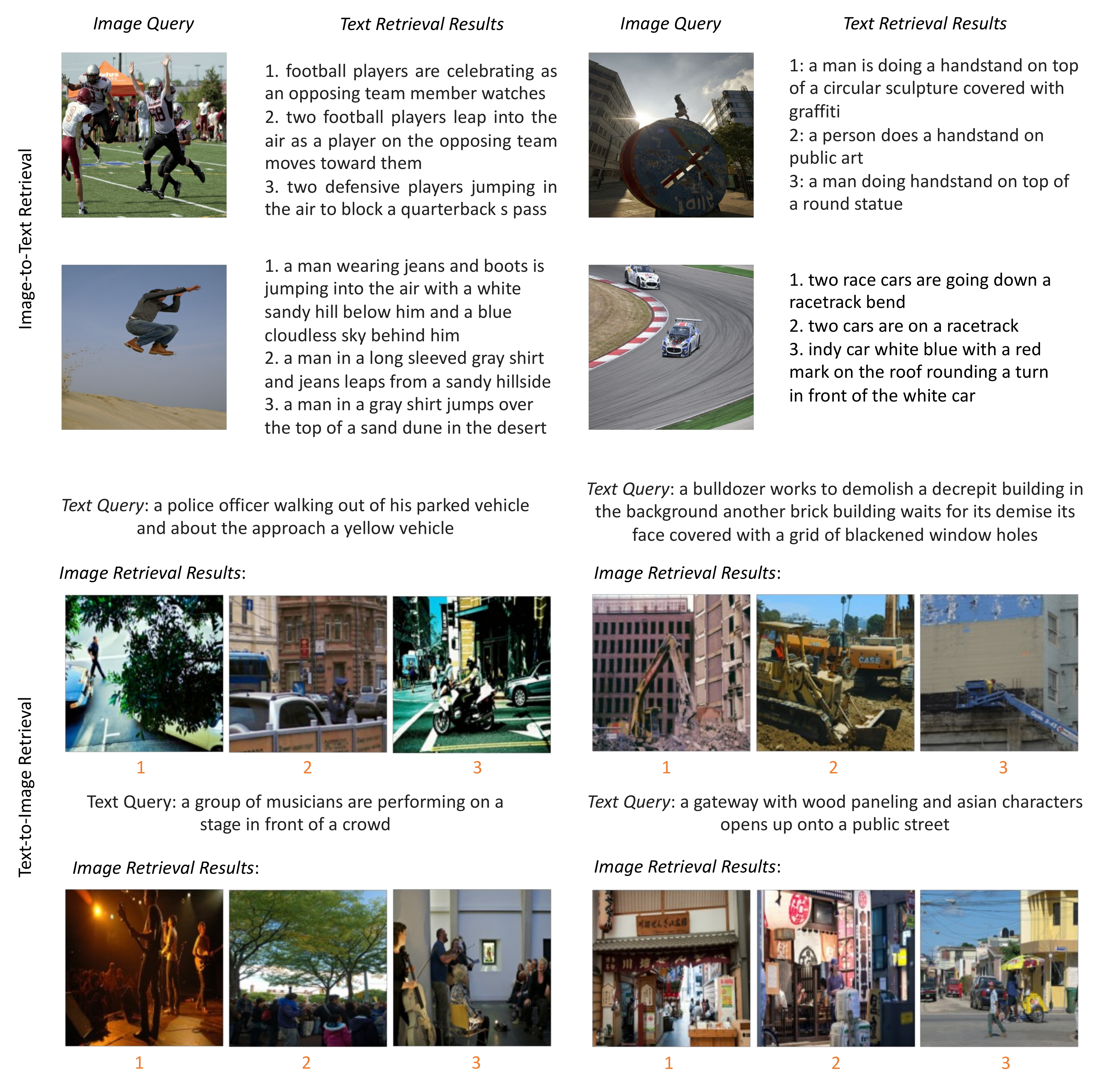}
    \vspace{-3mm}
    \caption{Visualization of image and text retrieval results. Despite training with orders of magnitude less resource, \ours correctly match many visual concepts with texts.}
    \label{fig:vis_retrieval_results}
    \vspace{-3mm}
\end{figure*}

\paragraph{Visualization of image and text retrieval.}\quad
We present image and text retrieval results of \ours in Fig.~\ref{fig:vis_retrieval_results}. Despite highly resource efficient training, \ours still produces accurate results on both image-to-text and text-to-image retrieval. For example, the concepts of football players, race cars, circular sculpture, police officer, musicians, and bulldozer are all correctly matched between image and texts.



\section{Conclusions}
We present the \ours, a method for resource-efficient language image pretraining. We propose to leverage small images with paired texts for the main constrastive training phase and finetune the model with high-resolution images for a short cycle at the end. The proposed training method has been validated on zero-shot image and text retrieval benchmarks and image classification datasets. In comparisons to the baseline method, \ours training recipe saves the computations by $6\sim8\times$ with improved zero-shot retrieval performance and competitive classification accuracy. Compared to the state-of-the art methods, \ours significantly saves $\bf79\%\sim98\%$ resource in cores$\times$hours with highly competitive zero-shot classification and image-text retrieval performance. We hope \ours paves the path to make contrastive language image pretraining more resource-friendly and accessible to the broad research community.

\subsubsection*{Broader Impact Statement}
Language image pretraining plays an important role in many applications, \eg image and text retrieval, text-to-image generations, open-vocabulary detection, etc. This work presents a language image pretraining method, \ours, on large-scale web datasets and the proposed model has been evaluated on a series of zero-shot downstream tasks. The large image-text corpus may contain biased or harmful content which could be learnt by the model. Our model is for research use only and \revise{these models should not be used in applications that involve detecting features related to humans (e.g. facial recognition).} The good news is \ours significantly reduces the resource use, thereby reducing the carbon footprint and is very environment-friendly for the community to build upon in the long run.




\bibliography{main}
\bibliographystyle{tmlr}


\end{document}